\documentclass[conference]{IEEEtran}
\IEEEoverridecommandlockouts
\usepackage{cite}
\usepackage{amsmath,amssymb,amsfonts}
\usepackage{algorithmic}
\usepackage{graphicx}
\usepackage{textcomp}
\usepackage{xcolor}
\usepackage{multicol}
\def\BibTeX{{\rm B\kern-.05em{\sc i\kern-.025em b}\kern-.08em
    T\kern-.1667em\lower.7ex\hbox{E}\kern-.125emX}}
    \makeatletter
 \let\old@ps@headings\ps@headings
 \let\old@ps@IEEEtitlepagestyle\ps@IEEEtitlepagestyle
 \def\confheader#1{%
 \def\ps@headings{%
 \old@ps@headings%
 \def\@oddhead{\strut\hfill#1\hfill\strut}%
 \def\@evenhead{\strut\hfill#1\hfill\strut}%
 }%
 \def\ps@IEEEtitlepagestyle{%
 \old@ps@IEEEtitlepagestyle%
 \def\@oddhead{\strut\hfill#1\hfill\strut}%
 \def\@evenhead{\strut\hfill#1\hfill\strut}%
 }%
 \ps@headings%
 }
 \makeatother

  \usepackage[pscoord]{eso-pic}
\newcommand{\placetextbox}[3]{
 \setbox0=\hbox{#3}
 \AddToShipoutPictureFG*{ \put(\LenToUnit{#1\paperwidth},\LenToUnit{#2\paperheight}){\vtop{{\null}\makebox[0pt][c]{#3}}}
 }
 }
 \placetextbox{.23}{0.055}{}
 
\begin{document}

\title{Comparative Study of Machine Learning Models and BERT on SQuAD\\}

\author{\IEEEauthorblockN{\textbf{Devshree Patel}\IEEEauthorrefmark{1},
\textbf{Ratnam Parikh}\IEEEauthorrefmark{1}, \textbf{Param Raval}\IEEEauthorrefmark{1} \textbf{and
Yesha Shastri}\IEEEauthorrefmark{1}}
\IEEEauthorblockA{School of Engineering and Applied Science,
Ahmedabad University\\
Email: [devshree.p,ratnam.p,param.r,yesha.s2]@ahduni.edu.in \\ \textbf{*All authors have contributed equally}
}}

\maketitle
\begin{abstract}
This study aims to provide a comparative analysis of performance of certain models popular in machine learning and the BERT model on the Stanford Question Answering Dataset (SQuAD). The analysis shows that the BERT model, which was once state-of-the-art on SQuAD, gives higher accuracy in comparison to other models. However, BERT requires a greater execution time even when only 100 samples are used. This shows that with increasing accuracy more amount of time is invested in training the data. Whereas in case of preliminary machine learning models, execution time for full data is lower but accuracy is compromised.
\end{abstract}

\begin{IEEEkeywords}
natural language processing, question answering, SQuAD, BERT
\end{IEEEkeywords}

\section{Introduction}
 As information in everyday life is increasing, it becomes difficult to retrieve a relevant piece of information efficiently. Thus, a Question-Answering (QA) system can be used to efficiently present the requested information. QA is an important application of NLP in real life which is a specific type of information retrieval method. The QA system makes an attempt to automatically find out the contextually and semantically correct answer for the provided question in text. Generally, the three components associated with the QA system are question classification, information retrieval, and answer extraction/generation.
 
 Though QA does not come without its challenges, one approach to get a machine to answer questions is Reading Comprehension (RC).Reading a text and answering from it is a challenging task for machines, requiring both understanding of natural language and knowledge about the world. The first step in the process to create such a system is a Question Answering dataset. A popular benchmark QA dataset created by Stanford University is known as the Stanford Question Answering Dataset or SQuAD. 
\section{Literature Review}
One of the first papers on applying machine learning approach for question answering tasks is Ng. et al., 2000. A semi-supervised learning approach for word representations was given by Turian et
al., 2010. A fast hierarchical language model was proposed by Mnih et al., 2009 which outperforms non-hierarchial neural model and best N-gram models. The task of QA is always challenging since it requires a comprehensive understanding of natural languages and the ability to do further inference and reasoning. In the original June, 2016 paper introducing SQuAD, Rajpurkar, et al [1]. developed a logistic regression model based on detailed linguistic features that achieved an F1 score of 51\%. Then, Xiong, et al. introduced a dynamic coattention network for encoding and attention modeling, Seo, et al. utilize bidirectional attention flow (BiDAF) encoding and attention schemes, combining question and context influence via a bidirectional LSTM. Later, using the self-attention based Transformer by Vaswani et. al.[2][4], Devlin, Jacob et. al. [3] from Google AI release the Bidirectional Encoder Representations and Transformer (BERT) model which surpassed the performance of all previous SOTA models on SQuAD 1.1.

BERT has demonstrated to have a good performance in question answering, sentence completion, and similar tasks when trained on a large corpus. Lan et al. [2020] proposes a lighter version of BERT called ALBERT [14] to address the issue of the model size growing too large on pretraining tasks and hurting overall performance. As of the time of writing of this paper, ensemble models with a variation of ALBERT paired with deep net architectures perform best on SQuAD 1.1 and SQuAD 2.0. To limit the scope of this paper, we perform a comparative study on SQuAD 1.1 and the original BERT model.
\section{Comparative Study}

\subsection{Data Pre-processing}\label{AA}
Stanford Question Answering Dataset (SQuAD) is a reading comprehension dataset which comprises of nearly 1,00,000 questions posed on several Wikipedia articles. This is a closed form dataset meaning that the answer to a question is always a part of the context and also a continuous span of context. For each observation in the training set, there is a context(paragraph), question and the answer text. Currently there are two versions of SQuAD available, but for simplicity of comparison we employ SQuAD v1.1. The goal is to predict the answer text for any new context and question provided. The above mentioned goal can be achieved by locating the correct sentence in the passage and then further finding the correct text in that sentence. 

We generate the sentence embeddings by using Facebook research’s InferSent[5] which is a method that provides semantic representations for English sentences. Being pretrained on a larger corpus and after building a preliminary vocabulary on SQuAD, we can get contextual embeddings with stronger weights to the more important/significant words of the sentence. By calculating the Euclidean distance and cosine similarity between sentences and questions, visualised in the multidimensional vector space, we can get an idea about the sentence that most closely gives the answer. See a diagrammatic explanation in the Appendix. Further, we attempted to use ML algorithms to determine the best fitting sentence embedding. The flowchart below shows the data preprocessing process.
\begin{figure}[t]
\centering
\includegraphics[scale=0.4]{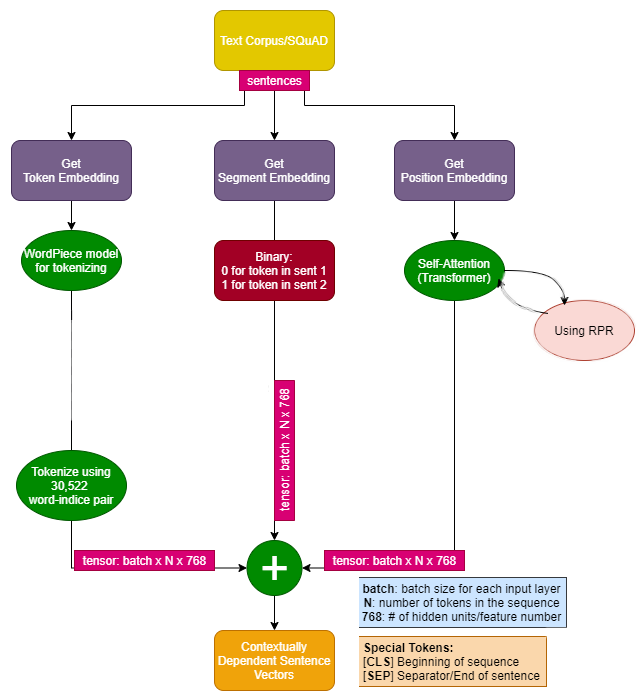} 
\caption{Diagram1}
\end{figure}

\subsection{Machine Learning Models}
The following are all the machine learning models applied on the SQuAD to show a detailed comparative study.\\
\subsubsection{\textbf{Logistic Regression}}
Logistic Regression is a supervised classification algorithm which is used to classify the predictions on the data into classes. Rajpurkar et al.[2016] showed that the performance of their logistic regression model was considerably low when compared with human performance. Their F1 score turned out to be 51\% for logistic regression while human performance F1 score was around 87\%. The f1 score calculated using our model turned out to be 52.5\% which is an improvement to the base model. Since the dataset was huge, we implemented the logistic regression model along with principal component analysis.Z-score normalisation was performed before computing the covariance matrix of the dataset.
After that, a projection matrix is formed using principal components corresponding to larger eigenvalues. For checking the effect of PCA on the data matrix, we again reconstructed it in order to calculate the reconstruction loss. We observed that on increasing number of principal components, reconstruction error reduces as now more number of components are used to span the dataset.

\subsubsection{\textbf{Gaussian Mixture Model}}
Gaussian Mixture Models are probabilistic models which can cluster the data points according to their probability distributions. Data points having a single distribution are grouped together. The clustering approach utilises both mean and variance and hence the results are more accurate. In order to minimize the cost function, Expectation-Maximisation(EM) Algorithm is used to compute the responsibility from the current parameters which is then used to update the parameters accordingly until the convergence. The data matrix formed after pre-processing was of \textit{85119x20} dimension. The 20 features include 10 columns of euclidean distance and 10 columns of cosine similarity which are normalised using Min-Max scalar. Since, the Gaussian distribution is plotted against 2 features, we select column 1 from the Euclidean distance and the corresponding column from the cosine similarities. We choose the number of components/clusters as 10.

\subsubsection{\textbf{Support Vector Machine}}
 SVM separates out different classes with margins for all the categories of classes. We have applied SVM on SQuAD with different kernels (linear, polynomial and radial basis function) and tuning the hyper-parameter values. The results were generated using optimal value of \textit{gamma}=0.1 (chosen from an array of different values like [0.0001, 0.001, 0.005, 0.1, 1, 3, 5]) and error limit of 1000. From all the kernels used above the \textit{RBF kernel} performs well on SQuAD with a training accuracy of 67\% and test accuracy of 66\%. SVM is known to be one of the best classification algorithm but does not create good enough representations of the sentences. 
\subsubsection{\textbf{Random Forest}}
The decision tree in random forest forms nodes to obtain a large number of data points from a class by finding values in the features which will divide the data samples in classes. The training and testing accuracy depends on the number of samples at leaf nodes and the number of trees. When the values of hyperparameters - \textit{min\_samples\_leaf}=8 and \textit{n\_estimators}=60,training accuracy is 77.8 percent and testing accuracy is 63.7 percent. However, when \textit{min\_samples\_leaf}=3 and \textit{n\_estimators}=5, training accuracy is 85.5 percent and testing accuracy is 58.7 percent. In general, as the value of both the parameters is decreased, training accuracy increases while testing accuracy decreases which denotes that model is overfitting the data.
\subsubsection{\textbf{XGBoost}}
The extreme gradient boosting mechanism helps improve performance as opposed to random forest because it works on trees with fewer splits. The training accuracy is 67\% and testing accuracy is 65\% which is lower than random forest. The testing accuracy will be more as we increase the value of \textit{max\_depth} parameter since it represents the depth of decision trees. But increasing it too much would result in over-fitting just as in Random Forest. For now, we keep \textit{max\_depth}=4. The default \textit{'deviance'} loss function works well for such textual and probabilistic outputs.

\begin{figure*}[t]
  \includegraphics[width=0.99\linewidth]{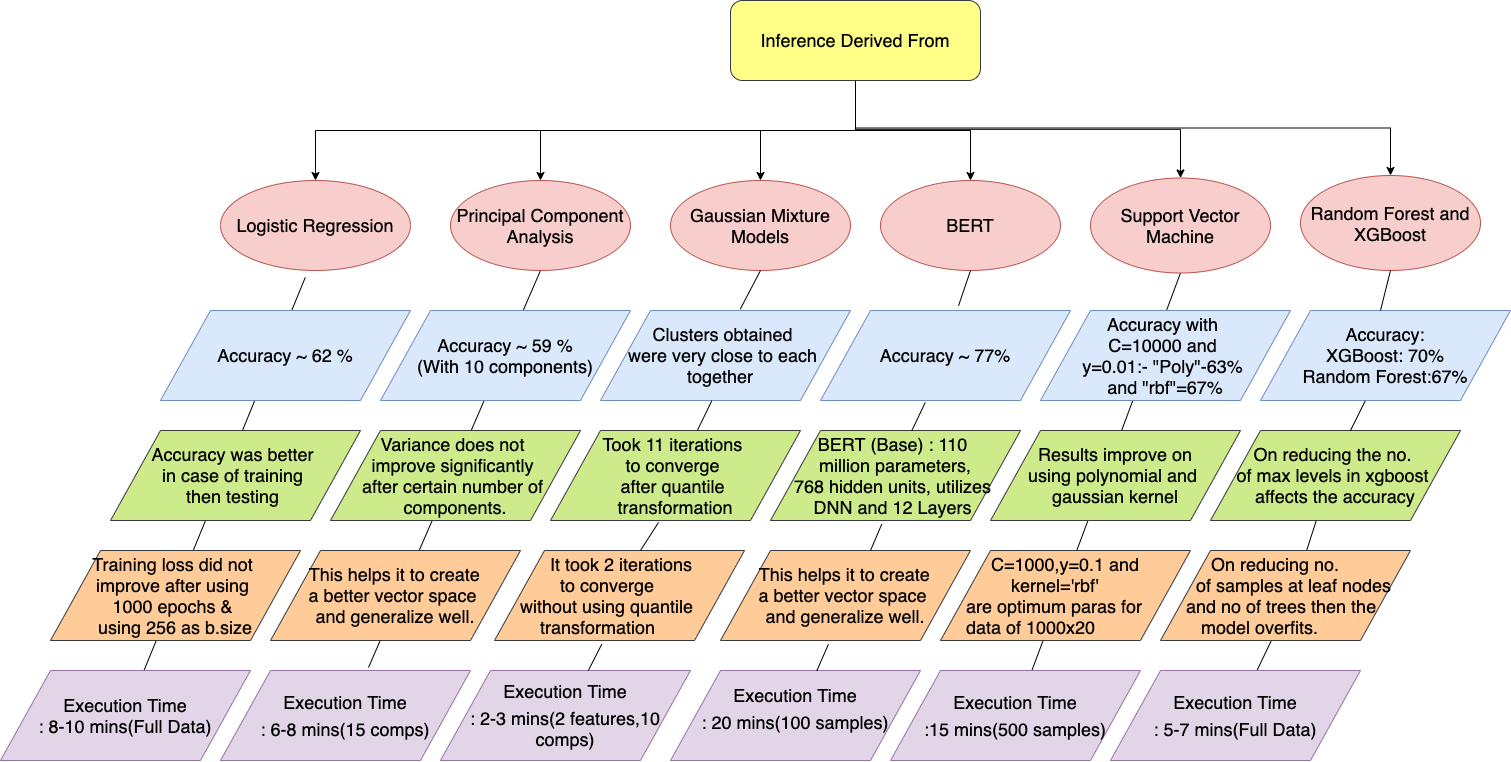}
  \caption{Inference Diagram}
\end{figure*}

\subsection{BERT}\label{AA}
In Devlin et al.[2019], the authors presented a language representation model called Bidirectional Encoder Representations from Transformers or BERT. During generation of sentence embedding vectors, BERT trains bidirectionally and looks at the whole sentence simultaneously in the way that humans look at the entire context of a sentence. A simple representation for understanding BERT is shown in \textit{Diagram 1}
Using, BERT$_{BASE}$ (L=12, H=768, A=12, Total Parameters=110M) pretrained and finetuned on SQuAD v1.1 on cloud TPU we get a test accuracy of 77\%.

\section{Results}
After applying various ML models and BERT to our dataset, we were able to derive the following inferences (as shown in the diagram below). Some models being too computational intensive for our PCs, we had to truncate our dataset to get a representative result.
\begin{itemize}
\item \textbf{Logistic Regression}\\
\textbf{Fig 3 and 4} shows the result obtained for logistic regression without regularisation. It can be observed that testing data accuracy starts little bit below than the training data whereas the loss for training and data curves are almost similar.
\begin{figure*}[!t]
    \centering
    \begin{minipage}[t]{0.48\linewidth}
        \centering
        \includegraphics[width=\linewidth]{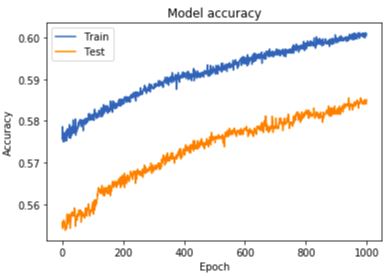}
        \caption{Accuracy}
    \end{minipage}
    \begin{minipage}[t]{0.48\linewidth}
        \centering
        \includegraphics[width=\linewidth]{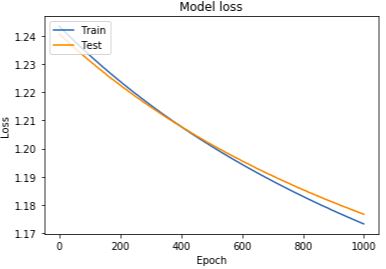}
        \caption{Loss}
    \end{minipage}
\end{figure*}
\begin{figure*}[!t]
\begin{center}
        \begin{minipage}[t]{0.48\linewidth}
        \centering
        \includegraphics[width=\linewidth]{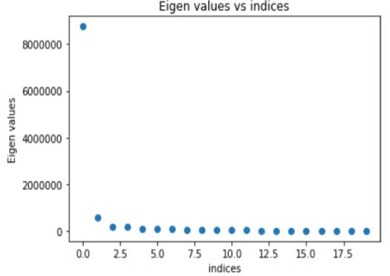}
        \caption{Eigen Values VS Indices}
    \end{minipage}
    \begin{minipage}[t]{0.5\linewidth}
        \centering
        \includegraphics[width=\linewidth]{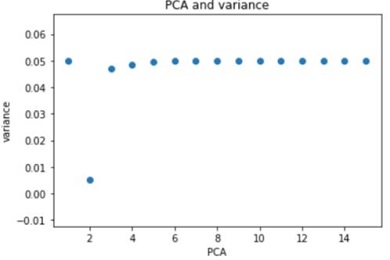}
        \caption{Principal Components VS Variance}
    \end{minipage}
    \begin{minipage}[t]{0.6\linewidth}
        \centering
       \includegraphics[width=\linewidth]{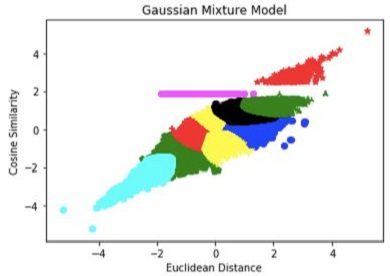}
        \caption{Gaussian Mixture Model Representation}
    \end{minipage}
    
\end{center}

\end{figure*}
\newline
\item  \textbf{Principal Component Analysis}\\
\textbf{Fig 5} represents trend between eigen values and its indices.The eigen values are sorted in descending order. In our case, the eigen values are nearly of similar values except the first eigen value.On arranging the eigen values in an unsorted manner, mean square error changes slightly.\\
\textbf{Fig 6} shows the trend between the principal components and the variance.The components represent the reduced version of euclidean distance and the cosine similarity between questions and the sentences of the passage which are similar in some cases.Thus,on increasing the number of components,our variance didn't change significantly because of similar components. \\
\item  \textbf{Gaussian Mixture Models}\\
\textbf{Fig 7} represents several clusters for different classified classes(10).The x-axis represents the euclidean distance and the y-axis represents the cosine similarity. Since, our total features were 20, we chose 2 from them- one column from euclidean and another from cosine similarity.The clusters obtained are closer too each other due to similarity between euclidean distances and cosine similarities between sentences of the passages.

\end{itemize}

\section{Conclusion}
We can conclude that despite using InferSent for generating contextual sentence representations, simple
ML models could not perform well (in most cases we observed over-fitting or poor performance despite increasing data samples or epochs). Comparing the results with a model like BERT, we can conclude that for comprehensive sentence representations larger and more complex models that employ DL techniques work better. Different versions of BERT such as BERT-Large, Multilingual-BERT, ALBERT (currently SOTA),RoBERTa, etc. can be studied to derive better inferences and results on SQuAD.

\onecolumn
\section{Appendix}
\subsection{Preprocessing the data}
\begin{figure}[h]
    \centering
    \includegraphics[scale=0.5]{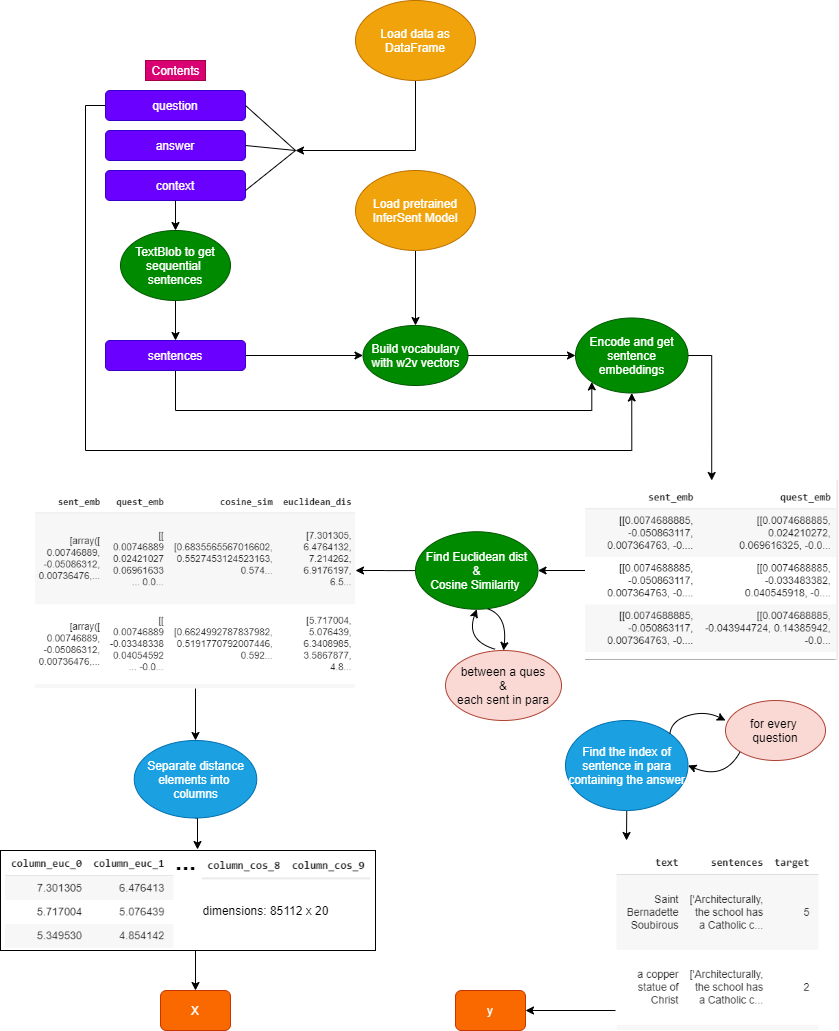}
    \caption{Preprocessing}
    \label{fig:my_label}
\end{figure}
\end{document}